\documentclass[letterpaper]{article} 
\usepackage[submission]{aaai24}  
\usepackage{times}  
\usepackage{helvet}  
\usepackage{courier}  
\usepackage[hyphens]{url}  
\usepackage{graphicx} 
\usepackage{float} 
\usepackage{parskip}  
\usepackage{natbib}  
\usepackage{caption} 
\usepackage{algorithm}
\usepackage{algpseudocode}
\usepackage{subfig}
\usepackage{graphicx}
\usepackage{amsmath}
 
\usepackage{amsthm}
\usepackage{tabularx}

\usepackage{cite}
\usepackage{amsmath,amssymb,amsfonts}
\usepackage{textcomp}
\usepackage{xspace}
\usepackage{makecell}
\usepackage{booktabs,siunitx}
\usepackage{diagbox}
\usepackage{array, multirow}
\usepackage{hyperref} 

\usepackage{newfloat}
\usepackage{listings}
\DeclareCaptionStyle{ruled}{labelfont=normalfont,labelsep=colon,strut=off} 
\floatstyle{ruled}
\newfloat{listing}{tb}{lst}{}
\floatname{listing}{Listing}
\title{Unaligning Everything: Or Aligning Any Text to Any Image in Multimodal Models}

\author{
    Shaeke Salman\textsuperscript{\rm 1}, 
    Md Montasir Bin Shams\textsuperscript{\rm 1}, 
    Xiuwen Liu\textsuperscript{\rm 1}
}
\affiliations {
    \textsuperscript{\rm 1}Department of Computer Science, Florida State University, FL 32306, USA\\
    \{salman, liux\}@cs.fsu.edu, mshams@fsu.edu 
}

\begin{document}

\maketitle

\begin{abstract}
Utilizing a shared embedding space, emerging multimodal models exhibit unprecedented
zero-shot capabilities. However, the shared embedding space could lead to
new vulnerabilities if different modalities can be misaligned. 
In this paper, we extend and utilize a recently developed effective gradient-based procedure that allows us to match the embedding of a given text by minimally modifying an image. Using the procedure,
we show that we can align the embeddings of 
distinguishable texts to any image through unnoticeable adversarial attacks in  joint image-text models, revealing that semantically unrelated images
can have embeddings of identical texts and at the same time
visually indistinguishable images can be matched to the embeddings of very different texts. Our technique achieves 100\% success rate when it is applied to 
text datasets and images from multiple sources.
Without overcoming the vulnerability,  multimodal models cannot robustly align inputs from different modalities in a semantically meaningful way. \textbf{Warning: the text data used in
this paper are toxic in nature and may be offensive to some readers.}
\end{abstract}

\section{Introduction}
Built on large pre-trained foundation models~\cite{bommasani2022}, 
applications have exhibited unprecedented capabilities for a wide range of tasks, setting new state-of-the-art on benchmark datasets, acing standard exams, and passing professional exams~\cite{openai2023gpt4, Protein2022Brandes, Usmle2023Kung, Law2023Chatgpt}. 
Such models, however, are not well understood
due to their complexity, even though the need for understanding and the risks of lacking is widely recognized and acknowledged~\cite{bommasani2022}.
For example, transformers have become a hallmark component in models for many applications and have led to significant improvements in performance on benchmark datasets~\cite{vaswani2023attention, dosovitskiy2021image, Devlin2018Bert}. 
By transforming inputs from different modalities (such as texts and images) 
to a common embedding space,
emerging multimodal models provide new capabilities and new applications are being developed by exploiting the shared space~\citep{radford2021learning}.

At the same time, it is well known that neural networks exhibit an intriguing property in that they are subject to adversarial attacks: some small changes to an input could result in substantial changes in model responses and outputs~\cite{goodfellow2015explaining,szegedy2014intriguing, chakraborty2018adversarial}. 
While studies have shown adversarial examples exist to break 
even aligned models~\cite{zou2023universal},
it is not clear whether the shared space could be exploited 
to establish arbitrary associations between images and texts,
or between two different modalities,
therefore breaking
the alignments that many of the models rely on in order to function properly.

\begin{figure*}[ht]
  \centering
  \includegraphics[width=0.95\textwidth]{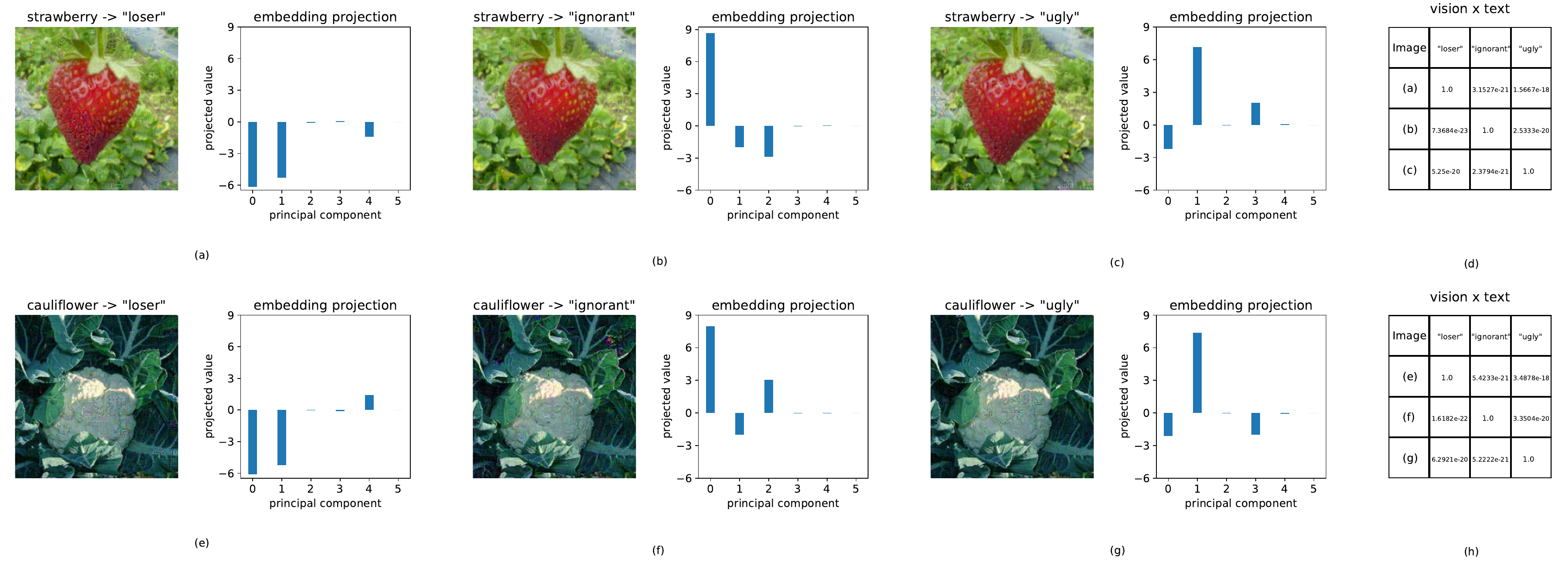}\label{fig:projection_embeddings}
  \caption{Typical examples from ImageNet obtained using the proposed framework. The visually indistinguishable images have different representations from each other as shown in their low-dimensional projections.  Note that the arrow in the title ($original \rightarrow target$) signifies a derived image from the original one by aligning the embedding of the original image with the target text embedding using our method. The projections of embedding-aligned images closely resemble the projections of the aligned text. The matrix shows the classification outcomes from the multimodal ImageBind pretrained model used directly with no modifications; each row corresponds to one image.
}
  \label{fig:overall}
\end{figure*}

In this paper, using a gradient-descent-based optimization procedure as detailed in our prior work~\cite{salman2024intriguing}, we show that perturbing an input image to a deployed model in unnoticeable ways can alter the resulting representation to match any chosen text and, therefore, reveal 
an inherent vulnerability of joint vision-text models. Since such multimodal models are being deployed,
the identified vulnerability should be considered for these models.
Furthermore, we show that the resulting inputs can dramatically change classification results with no modifications to the classifiers. 

To highlight the main advantages of our framework, we present our results using multiple models, including the ImageBind~\cite{girdhar2023imagebind}. Fig. \ref{fig:overall} shows several images along with their representations and the classification results. The three visually indistinguishable images in the top row of Fig. \ref{fig:overall} (see Fig. \ref{fig:diffnoise} in
appendix for pixel differences) have very different representations, as shown by their low-dimensional projections; the images in the bottom row also have very different representations.
On the other hand, the pairs in each of the three columns in Fig. \ref{fig:overall}  have very similar representations even though they are semantically very different.
When we pass these images to the unmodified multimodal ImageBind model, the images with similar embeddings are classified into the same class, regardless of their semantic similarity, as shown in Fig. \ref{fig:overall} (d) and (h). 

These and additional results shown in the Experiments section, along with the fact we have obtained the same findings on all the image-text pairs we have used, demonstrate convincingly that there are visually indistinguishable inputs corresponding
to the embeddings of very different texts, and yet there are very different images corresponding
to the embeddings of identical texts. 
By analyzing the equivalence classes~\cite{salman2024intriguing} of the embeddings of vision-text models, the vulnerability we show is 
due to the representations used by such models and do not depend on application-specific classifiers.

Our main contributions are as follows:

\begin{itemize}
  \item  Using our efficient computational procedure to match specified representations, we clearly show an inherent vulnerability of joint vision-text models,
  where arbitrary associations between the images and texts can be established, regardless of the semantics of the images and texts. More specifically, we show that visually indistinguishable images can have very different representations, yet 
  unrelated images semantically can correspond to similar text embeddings.
   \item We show that we can map all the texts in different datasets to visually indistinguishable images with a 100\% success rate.
\end{itemize}



\section{Related Work}

The transformer architecture \cite{vaswani2023attention} revolutionized NLP by effectively capturing long-range dependencies, resulting in powerful pre-trained models like BERT \cite{Devlin2018Bert} and GPT \cite{brown2020language} that excel in various tasks. This advancement extends to computer vision with the Vision Transformer (ViT) \cite{dosovitskiy2021image}, showcasing the transformative impact of attention mechanisms across domains.

The recent prompting-based models and multimodal models have further accelerated the trend. The joint multimodal models have demonstrated significant benefits by employing a shared embedding space across various modalities. For example, CLIP \cite{radford2021learning}
aligns vision and text representations. The model is trained to predict the coherence of image-text pairs, which enables it to understand complex relationships between the two modalities.  Several recent works extend the technique of shared embedding space beyond text and vision by
employing a unified embeddings space in various modalities are: GPT-4 \cite{openai2023gpt4}, MiniGPT-4 \cite{zhu2023minigpt}, Flamingo \cite{alayrac2022flamingo}, Bard \cite{Bard}, LLaVA \cite{llava} and, ImageBind \cite{girdhar2023imagebind}.
Another line of research aims to comprehend models by probing them to unveil new properties. A well-studied problem is adversarial attacks, where unnoticeable changes to the input can cause the models, primarily classifiers, to change their predictions. Most adversarial attacks are applied
to commonly used (deep) neural networks, including multiple-layer perceptrons and
convolutional neural networks (CNNs), demonstrating
their vulnerability and sensitivities to such adversarial changes~\cite{szegedy2014intriguing,goodfellow2015explaining}. Croce and Hein propose AutoAttack, 
an ensemble of parameter-free attacks that combines multiple methods to provide a robust assessment of a model's vulnerability \cite{croce2020reliable}. Recent studies have explored the vulnerability of multimodal models to adversarial attacks, which can potentially jailbreak aligned large language models (LLMs) or Vision Language models (VLMs) \cite{carlini2023aligned,qi2023visual,zou2023universal}. 
Bhojanapalli et al. investigate the robustness of ViTs against attacks where the attacker has access to the model's internal structure \cite{bhojanapalli2021understanding,shao2022adversarial}. Notably, these methods revolve around generating adversarial examples based on the classifier's methodology rather than focusing on the representation level. However, our approach is different. Rather than crafting an adversarial example tailored to a particular classifier, our method is designed to generate examples that conform to a specified representation.

A closely related study by Kazemi et al. examines the behavior and vulnerabilities of the CLIP model \cite{kazemi2024learn}. Their work is centered on inverting the CLIP model embeddings to understand the semantic information of these embeddings. However, our work focuses on demonstrating the vulnerabilities within the shared embedding spaces of multimodal models through adversarial attacks. We show through extensive experiments that visually indistinguishable images can be mapped to arbitrary text,
revealing an inherent vulnerability that is classifier agnostic. 

\section{Preliminaries}

As this paper focuses on vision-language models that are based on transformers, here we first describe the transformers mathematically and then describe the vision-language models. Transformers can be described mathematically succinctly, consisting of a stack of transformer blocks.
A {transformer block} is a parameterized function class $f_\theta: \mathbb{R}^{n \times d} \rightarrow \mathbb{R}^{n \times d}$. If $\mathbf{x} \in \mathbb{R}^{n \times d}$ then $f_\theta(\mathbf{x}) = \mathbf{z}$ where
$Q^{\left(h\right)}\left(\mathbf{x_i}\right) = W^T_{h,q}\mathbf{x}_i,\quad
    K^{\left(h\right)}\left(\mathbf{x_i}\right) = W^T_{h,k}\mathbf{x}_i,\quad
    V^{\left(h\right)}\left(\mathbf{x_i}\right) = W^T_{h,v}\mathbf{x}_i,\quad
    W_{h,q}, W_{h,k}, W_{h,v} \in \mathbb{R}^{d \times k}$.
The key multi-head self-attention is a softmax function applying row-wise on the inner products.\footnote{Note that there are other ways to compute the attention weights.}
\begin{equation}
    \alpha_{i,j}^{\left(h\right)} = \texttt{softmax}_j\left(\frac{\left<Q^{\left(h\right)}\left(\mathbf{x}_i\right),K^{\left(h\right)}\left(\mathbf{x}_j\right)\right>}{\sqrt{k}}\right).
\end{equation}
The outputs from the softmax are used as weights to compute new features, emphasizing the ones with higher weights given by
\begin{equation}
    \mathbf{u}'_i = \sum\limits_{h=1}^H W^T_{c,h} \sum\limits_{j=1}^n \alpha_{i,j} V^{\left(h\right)} \left(\mathbf{x}_j\right),\quad
    W_{c,h} \in \mathbb{R}^{k \times d}.
\end{equation}

The new features then pass through a layer normalization, followed by a ReLU layer, and then another layer normalization. 
Typically transformer layers are stacked to form deep models. 

\begin{figure}[ht]
  \centering
  {\includegraphics[width=0.46\textwidth]{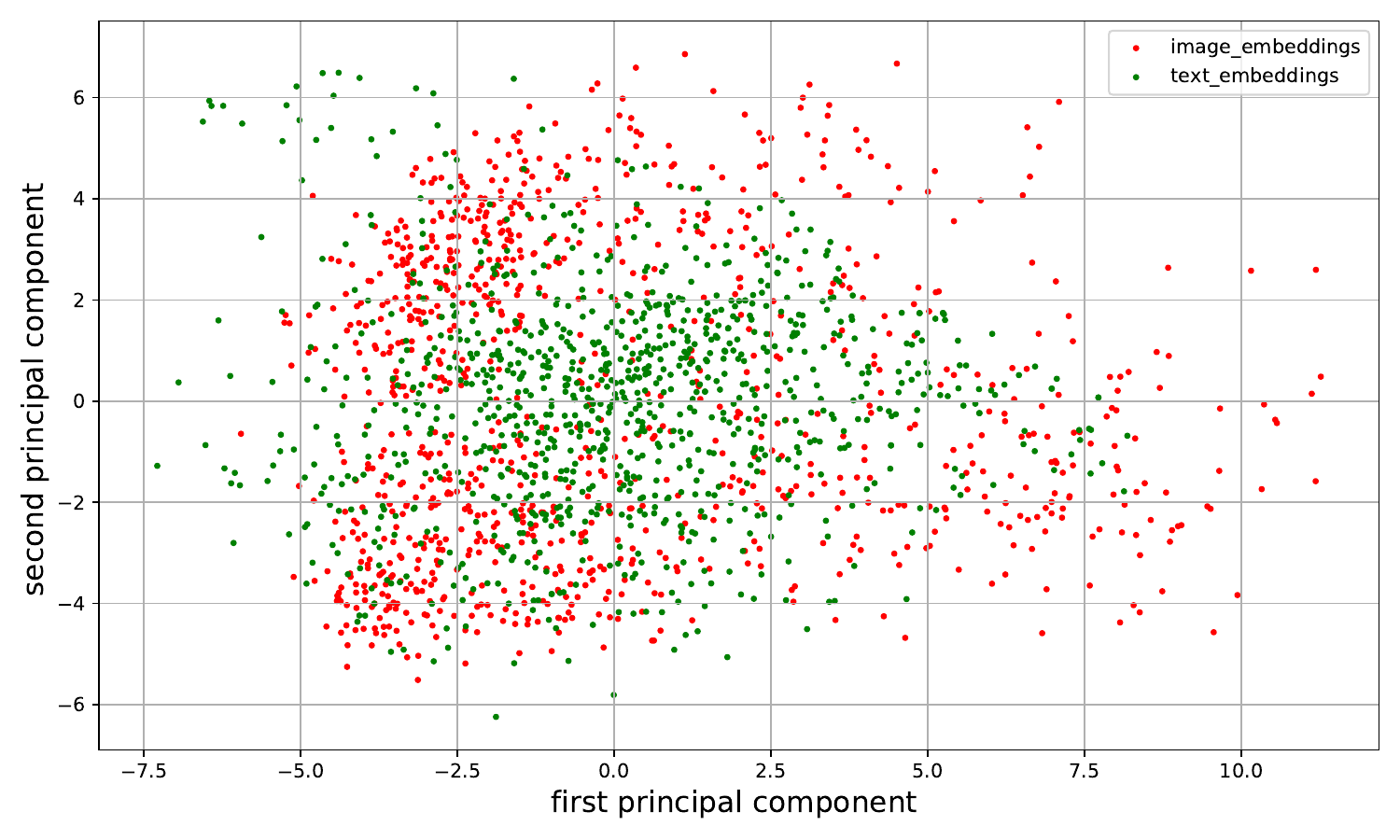}}
  \caption{Low-dimensional projections of the embeddings of images and texts, showing texts and images share the same embedding space.  We use all of the toxic comments (i.e., 992) from the 1,2,3-tokens toxic dataset and the same number of strawberry and cauliflower images from ImageNet.
  }
  \label{fig:proj_2d}
\end{figure}

While transformer models are widely used for natural language processing tasks, recently, they are adapted to vision tasks by using image blocks on the basic units, and spatial relationships among the units are captured via the self-attention mechanism~\citep[]{dosovitskiy2021image}. A vision-language model based on transformers incorporates a dedicated transformer model for each input modality. The resulting representations from these modalities are mapped to a shared embedding space.

For the ImageBind model, we denote the model for image $x$ as $f_I(x)$ and for text $t$ as $f_T(t)$. 
Fig. \ref{fig:proj_2d} shows that the image embeddings and text embeddings share 
the same vector space. 
Given image $x$ and $C$ text labels, $t_0, \ldots, t_{C-1}$, the zero-shot classification uses softmax applied on the dot products of the image and text representations. Therefore, the probability classified $x$ to $t_i$ is given by
\begin{equation}
\frac{e^{f_I(x)^T f_T(t_i)}}{\sum_{j=0}^{C-1}e^{f_I(x)^T f_T(t_j)}},
\end{equation}
which is a typical implementation of the softmax function. Note that the probabilities reported in this paper's figures are computed using a publicly available ImageBind model without any change. While the proposed method applies to all transformer-based models with continuous inputs, we focus on multiple models, including the CLIP model~\citep[]{radford2021learning}, which jointly models images and text using the same shared embedding space as the ImageBind \cite{girdhar2023imagebind} model. Ideally, only images and texts that are semantically related
should have similar embeddings. The image and text embeddings can help each
other, resulting in robust zero-shot capability~\citep{radford2021learning}. 
On the other hand, vulnerabilities in associations between images and texts could be exploited, resulting in new weaknesses.

\section{Methodology }
Understanding the structures of the representation space is crucial for determining how the model generalizes. As introduced in our earlier work~\cite{salman2024intriguing}, we have proposed a framework to explore and analyze the embedding space of vision transformers, uncovering intriguing equivalence structures and their implications for model generalization and robustness. 
Generally, we model the representation given by a (deep) neural network (including a transformer) as a function $f: \mathbb{R}^m\rightarrow\mathbb{R}^n$. A fundamental question is to have a computationally efficient and effective way to explore the embeddings of inputs by finding the inputs whose representation will match the one given by $f(x_{tg})$, where $x_{tg}$ is an input whose embedding we like to match. Informally, given an image of a strawberry in Fig.~\ref{fig:overall} as an example, all the images that share its representation given by a model will be treated as a strawberry. 

\subsection{Embedding Alignment Procedure}
 As described in our previous work~\cite{salman2024intriguing}, the proposed approach for embedding alignment focuses on aligning the representation of an input with that of a target input. The core of the method is an iterative gradient optimization procedure, similar to how most of the neural networks are trained, except the gradient is calculated with respect to the input variables. Since we need to match two vectors, we define the loss for finding an input matching a given representation as
\begin{equation}
   L(x) = L(x_0+\Delta x)= \frac{1}{2}\Vert f_I(x_0+\Delta x) - f_T(t_{tg})\Vert^2,
\end{equation}
where $x_0$ is an initial image and $f_T(t_{tg})$ specifies the target embedding for a specified text sequence $t_{tg}$. Approximately, the gradient is given by
\begin{equation}
\frac{\partial L}{\partial x} \approx \left(\frac{\partial f}{\partial x}\Big|_{x=x_0}\right)^T(f_I(x_0+\Delta x) - f_T(t_{tg})).
\label{eq:grad_J}
\end{equation}

In each step, the algorithm first calculates the loss as defined by the loss function, between the image embedding with the target embedding as the one given by the specified text. Then using PyTorch, it computes the gradient by doing backward computation. After the gradient is calculated, we update the pixel values by doing gradient descent.

\begin{figure*}[ht]
    \centering
      \includegraphics[width=0.90\textwidth]{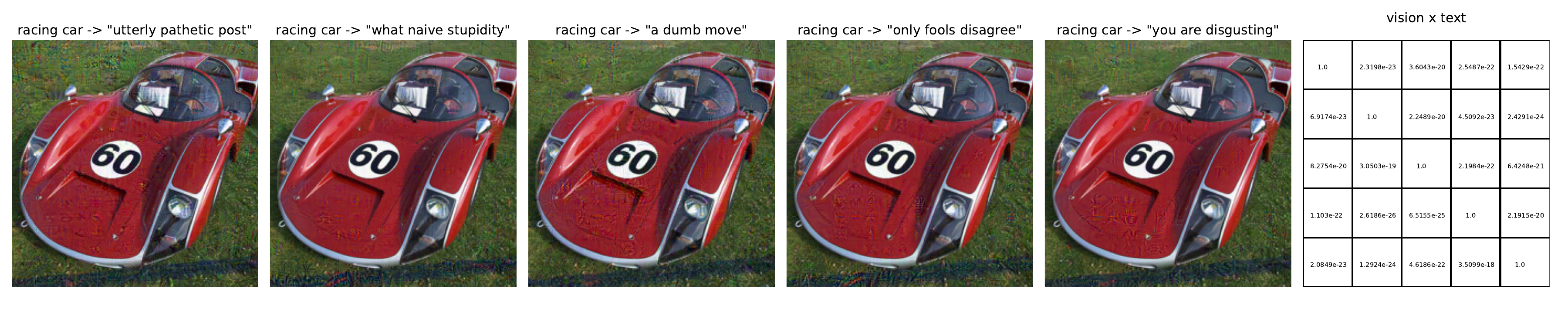}\label{fig:more1}
      \includegraphics[width=0.90\textwidth]{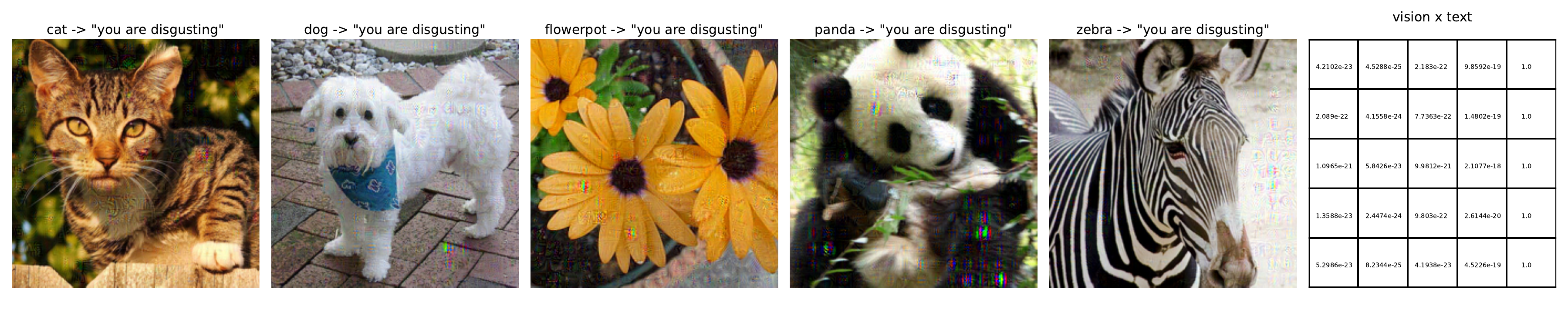}\label{fig:more2}
      
    \caption{
    (top) More examples involving ImageNet and 1,2,3-tokens toxic dataset, where visually indistinguishable images have very different representations via embedding alignment with the corresponding texts and therefore very different classification outcomes (as shown in the classification probabilities; each row in the matrix corresponds to one image (from left to right)). (bottom) Visually very different images have very similar embeddings, aligned and classified to a particular text.
    }
 \label{fig:more_examples}
\end{figure*}

\begin{figure}[ht]
  \centering
  {\includegraphics[width=0.23\textwidth]{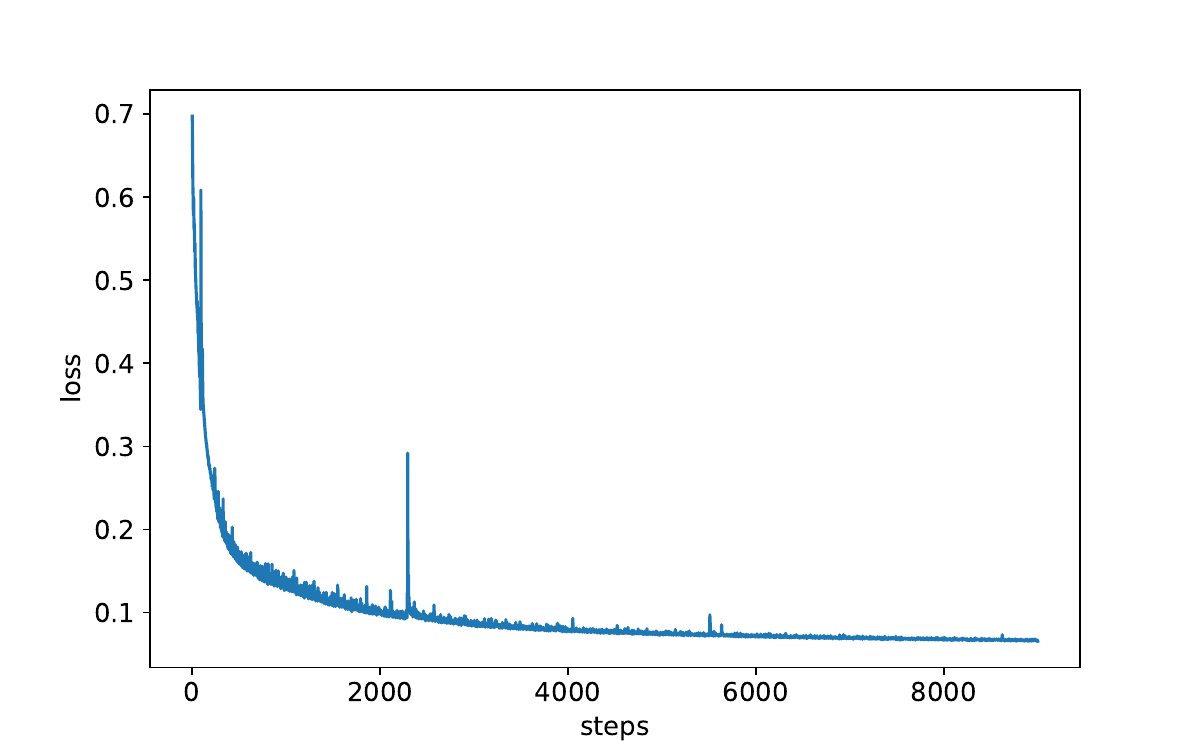}}
  {\includegraphics[width=0.23\textwidth]{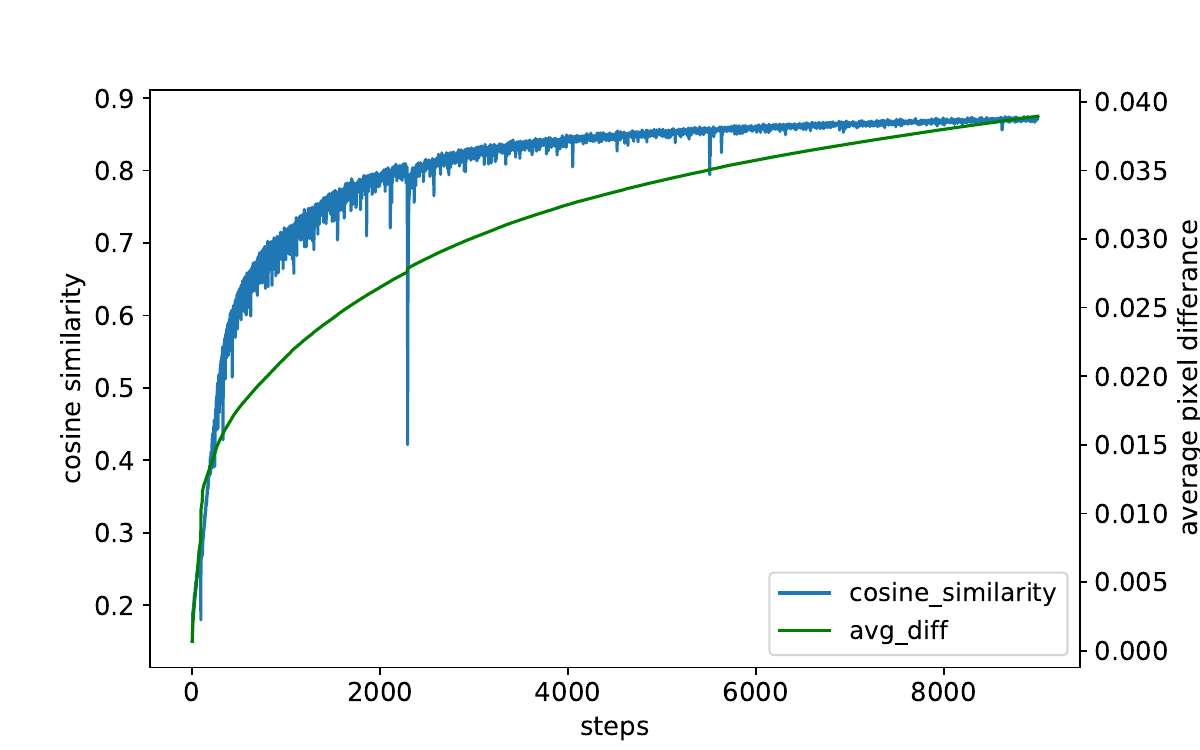}}
  \caption{The evolution of loss while matching a target embedding. (left) the loss w.r.t. steps. (right) the cosine similarity between the embeddings of the new input and the target w.r.t. the steps, along with the average pixel value difference between the new input and the original image.}
  \label{fig:training_dynamics}
\end{figure}

\begin{figure}[ht]
  \centering
  {\includegraphics[width=0.47\textwidth]{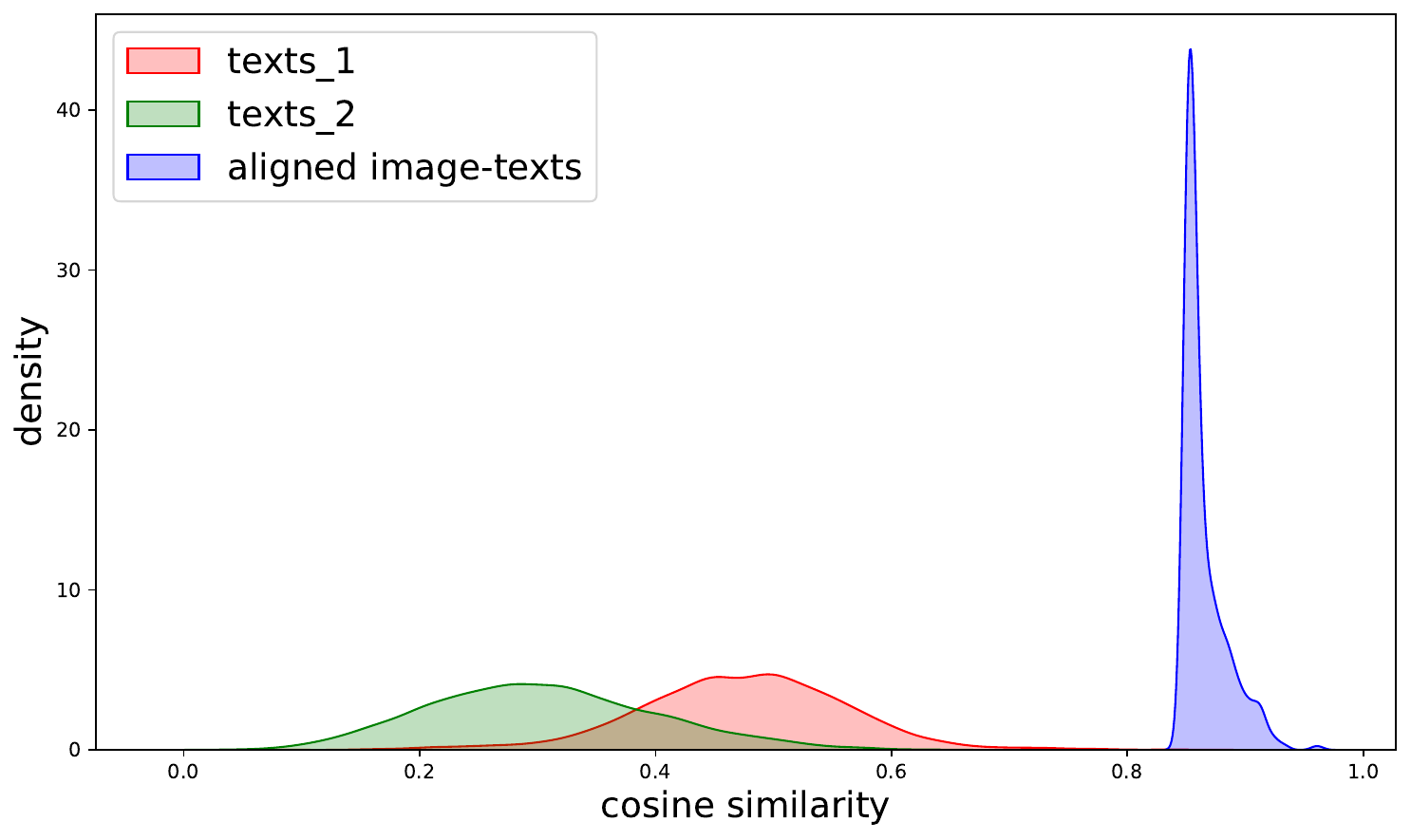}}
  \caption{(to be viewed in color) Cosine similarity distribution. The red and green ones stand for the cosine similarity values corresponding to pairs of texts (i.e., embeddings) from the two toxic datasets considered. The blue one shows the distribution of cosine similarities of the embeddings of embedding-aligned image and text pair from the ImageNet and toxic dataset. As the cosine similarities of toxic data pairs do not overlap with other embeddings, potential mapping opportunities exist. 
  }
  \label{fig:cosine_sim}
\end{figure}

\begin{figure*}[ht]
  \centering
  \includegraphics[width=0.95\textwidth]{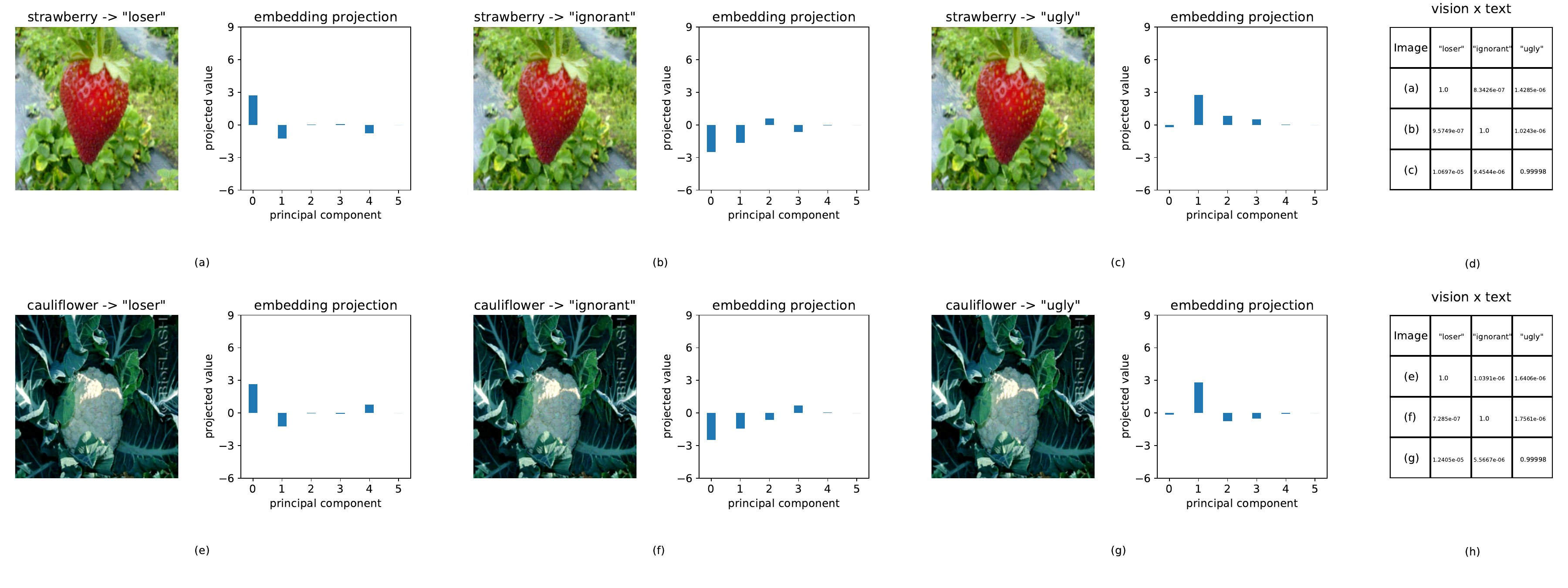}\label{fig:projection_embeddings_2}
  \caption{
  Examples obtained using the proposed framework
for different multimodal models, such as CLIPSeg. The results are given in the same format as depicted in Fig. \ref{fig:overall}. The example demonstrates that the method is model-agnostic.
}
  \label{fig:overall_2}
\end{figure*}

\begin{figure*}[ht]
  \centering
  \includegraphics[width=0.95\textwidth]{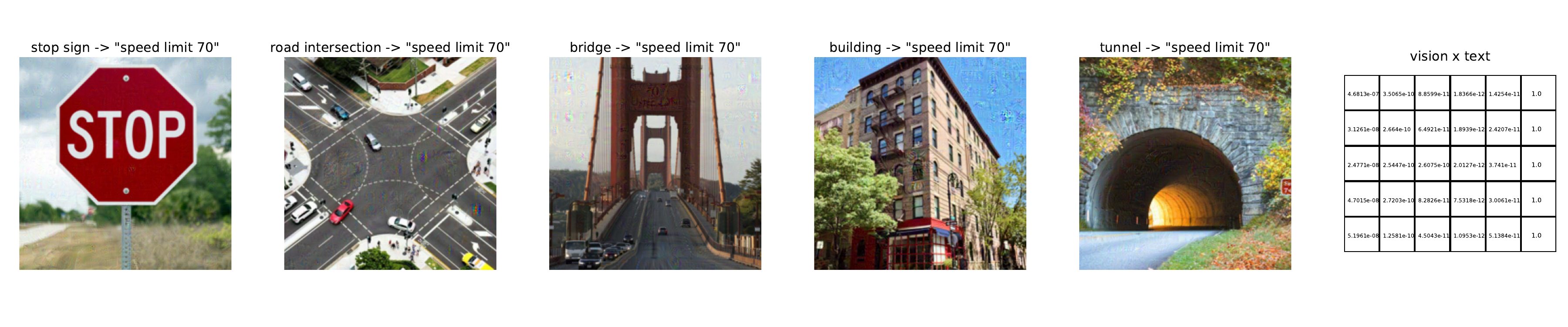}\label{fig:more_rl}
  \caption{Real-world scenarios where the proposed method is applied. The images are taken randomly from the web. All the matched images (for a stop sign, a road intersection, a
bridge, a building, and a tunnel) are recognized as the
sign ``speed limit 70". The examples demonstrate that our method works robustly for any data examples, therefore, the method is dataset-agnostic.
}
  \label{fig:more_realw}
\end{figure*}

Eq.~\eqref{eq:grad_J} shows how the gradient of the mean square loss function is related to the Jacobian of the representation function at $x=x_0$. 
In other words, the gradient is related to the differences between the current and 
target text representations. When the differences are large, the gradient should be significant as well. Consistent with this analysis, on all the examples we have experimented with, we are able to minimize the loss, and details are provided in the Experimental Results section and appendix.

While local optimal solutions could be obtained by solving a quadratic programming problem or linear programming problem, depending on the norm to be used when minimizing $\Delta x$, the gradient function works effectively for all the cases we have tested due to the Jacobian of the transformer.

One of the practical issues using the gradient descent-based procedure is how to determine the learning rate. In the case of the transformers, the model can be approximated by a linear model when it moves within one activation region; note that it is approximate due to the nonlinearity of the softmax. The algorithm 
is able to find the matching representations for a wide range of learning rates; see the Experimental Results section for more details.

\section{Experiments}
In this section, we first outline the specifics of our experimental settings and implementation details. Our designed framework is systematically applied across various datasets and multiple multimodal models; in the subsequent subsections, we present both the experimental outcomes and quantitative results. Our findings showcase the capability to align any distinguishable text with an image through imperceptible adversarial attacks within a joint image-text model. More importantly, we show that our framework exhibits versatility, being agnostic to both the model architecture and dataset characteristics.

\subsection{Datasets and Settings}
 
\textbf{Datasets.} We conduct extensive experiments to evaluate our proposed framework on widely recognized publicly available vision datasets, namely ImageNet~\cite{imagenet2009Deng} and MS-COCO~\cite{lin2015microsoft}. For the text aspect, we adopt a methodology inspired by the work of \citet{jones2023automatically} and \citet{borkan2019nuanced}, obtaining a dataset comprising 68, 332, and 592 toxic comments with 1, 2, and 3 tokens respectively\footnotemark\footnotetext{https://github.com/ejones313/auditing-llms/tree/main/data}. Additionally, our framework undergoes evaluation using the Jigsaw toxic dataset available at Kaggle\footnotemark\footnotetext{https://www.kaggle.com/c/jigsaw-toxic-comment-classification-challenge}~\cite{vanaken2018challenges}.

\textbf{Implementation Details.} To demonstrate the feasibility of the proposed method on large multimodal models, we have used the pretrained model publicly available by ImageBind\footnotemark\footnotetext{https://github.com/facebookresearch/ImageBind}, which in turn uses a CLIP model.\footnotemark\footnotetext{https://github.com/mlfoundations/open\_clip} 
More specifically, ImageBind utilizes the pre-trained vision (ViT-H 630M params) and text encoders (302M params) from the OpenCLIP~\cite{ilharco_gabriel_2021_5143773, girdhar2023imagebind}. The input size is $224\times224\times3$, and the dimension of the embedding is 1024. We perform all our experiments on a lab workstation featuring two NVIDIA A5000 GPUs. We will provide source code for all our experiments in GitHub\footnotemark\footnotetext{https://github.com/programminglove08/UnalignMM}.

\textbf{Additional Models.} To demonstrate the broader applicability of the models, we thoroughly evaluate with several other multimodal models, including CLIPSeg~\cite{lüddecke2022image}, AltCLIP~\cite{chen2022altclip}, BLIP-2~\cite{li2023blip2}, etc. An example with CLIPSeg is presented in the following subsection.
The size of the input is determined by the models. For all the multimodal models we have used, a preprocessing step is used to resize the input image to 224 × 224 × 3 for subsequent processing. Therefore, the method works equally well regardless of the resolution of the original input images.

\textbf{Embedding Projections.} To obtain the low-dimensional projections of the images shown in Fig. \ref{fig:overall} and other similar figures, the largest principal components are computed from a subset of images from the ImageNet dataset. Then, we project an embedding to be displayed along the six principal components with the largest eigenvalues. Note that the projections are used to illustrate the differences between embeddings, and details of the principal components would not impact the results significantly in that similar embeddings will have similar projections, and different embeddings will have different projections.

\subsection{Experimental Results}
To demonstrate the effectiveness of our method on deployed models such as the CLIP model,
a key is to be able to match a given representation given by a phrase, a sentence, or any text sequence that can be
encoded by the text transformer. 
We have tested the embedding matching procedure using many image and text pairs and
Fig. \ref{fig:training_dynamics} shows a typical example. 
The left of Fig. \ref{fig:training_dynamics}  shows the evolution of the loss when matching the embedding of an image to a specified target embedding. Similar to gradient descent, where the loss could become higher or lower, the high peaks indicate noise during the optimization process because the loss is non-linear. We use a small step size to make sure it converges. The right panel shows that cosine similarity increases steadily. We also show the average pixel value difference between the new input and the original image at each step; one can see the values remain very small even though they increase as well. The algorithm is not sensitive to the learning rate and works effectively across a broad range of values, spanning from 0.001 to 0.09. For instance, with a learning rate of 0.001, convergence is achieved in around 40,000 iterations, while 0.09 requires around 8,000 iterations. The visual differences in the resulting images are not noticeable. Eqn. 4 and 5 provide an explanation, as the gradient for our loss is insensitive to the learning rate. 

\textbf{Systematic evaluation. } To further demonstrate the effectiveness of the gradient procedure to match embeddings, we have applied them to numerous images and texts from different sources. Understanding the algebraic and geometric structures of the embedding space allows us to explore the space effectively. For example, we can find adversarial attacks to the embedding of any given image or text using the proposed gradient procedure. Fig. \ref{fig:overall} shows two examples. To demonstrate the universal applicability of the procedure and the adversarial examples that exist almost everywhere, Fig. \ref{fig:more_examples} shows more examples from different categories from the ImageNet dataset. See the appendix for additional examples on ImageNet and MS-COCO datasets. The efficacy of the procedure is model-agnostic. To substantiate and confirm this, Fig. \ref{fig:overall_2} illustrates an example of a different multimodal model using CLIPSeg, showcasing the consistent application and effectiveness of the approach irrespective of the specific model employed. In addition, Fig. \ref{fig:more_realw} shows real-world scenarios, demonstrating the practical relevance of our findings. All these examples convincingly demonstrate our method is model and dataset-agnostic.

\begin{table}
  \centering
    \small
  \begin{tabular}{lll}
    \toprule
    Data & Match Success Rate & Mean $\ell_2$ Distortion \\
    \midrule
    1-Token & 100\% & 0.98 $\pm$ 0.09  \\
    2-Token & 100\% & 0.83 $\pm$ 0.15 \\
    3-Token & 100\% & 0.47 $\pm$ 0.11 \\
    \bottomrule
  \end{tabular}
  \label{tab:match_success_rate}
  \caption{Success rates and mean $\ell_2$ distortions after we align the embeddings of given images (from ImageNet)
  to all the 1, 2, and 3-token toxic comments, respectively, in the toxic dataset. 
  }
\end{table}

\textbf{Quantitative evaluation.} Fig. \ref{fig:cosine_sim} depicts the distribution of cosine similarities: the red and green curves represent cosine similarity values corresponding to pairs of text embeddings from the two toxic datasets under consideration respectively. The blue curve illustrates the distribution of cosine similarities between embeddings of image and text pairs aligned through our embedding process from the ImageNet and toxic dataset. 
Essentially, the absence of overlap indicates that we can subtly modify an image corresponding to any selected text. In other words, with our approach, if we are provided with two or more texts, we can generate multiple visually indistinguishable  images, one  for each text, ensuring that a classifier will  classify every image to the assigned text, regardless of the semantics of the images. Due to this characteristic, Table 1 demonstrates a $100\%$ success rate~\cite{carlini2023aligned} in accurately matching the images with the toxic texts. To define the success rate, we first establish criteria for a successful image alignment. After aligning an image with the embedding of a specific text, we utilize the imageBind model for classification. If the resulting classification matches the given text, the alignment is considered successful; otherwise, it is not. The success rate on a particular dataset is then calculated as the percentage of images that meet these criteria. As a concrete example, Fig. \ref{fig:unclipped_68} illustrates the alignment of each of the 68 different text embeddings with an image. Since all cases are successful, the success rate is 100\%.
In addition to metrics such as attack success rate and mean $\ell_2$ distortion shown in Table 1, we note the number of pixels changed above specified thresholds ($>0.03 : 4500 | >0.05 : 795 | >0.08 : 90 | >0.1 : 25 | >0.2 : 1$) and the mean $\ell_\infty$ norm of the difference images (0.09, 0.07, and 0.03 for 1, 2, and 3-token comments, respectively), as these additional evaluation metrics are commonly used.

\textbf{Adversarial Modification Detection:} We have observed that the embedding-matched images exhibit much higher sensitivity to Gaussian noise than the original ones. Leveraging this insight, we have designed a detection algorithm introduced in our previous work~\cite{salman2024zshot}. As demonstrated in that study, the detection algorithm performs reliably and consistently across a wide range of standard deviations. The process is as follows: we add Gaussian noise of a specified standard deviation to a given image and then classify them. If the labels of the two images agree, the image is unmodified; otherwise, the image is modified. 


\section{Discussion and Future Work}


It may be attempting to categorize our framework as an adversarial attack technique. Our primary focus is on analyzing the embedding space; we utilize the ImageBind solely as a classifier to validate our findings and is not used otherwise. While our embedding matching procedure can be used to generate effective adversarial examples, it is fundamentally different. Our technique is classifier agnostic and does not exploit features specific to classifiers. Consequently, our examples with matched embeddings will appear to be the same to any classifier or downstream model that builds on embeddings. On the other hand, traditional adversarial attacks are specific to classifiers and applications, focusing on altering their outputs by changing the input.


Identifying adversarial attacks on multimodal models is very active~\cite{qi2023visual,schlarmann2023adversarial,evtimov2021adversarial}. In general, all of them focus on how small changes in inputs can alter the final output (such as captions or classification labels) across various models. In contrast, our work identifies a new representation vulnerability. For instance, as shown in Fig. \ref{fig:overall}, three strawberry and cauliflower pairs can be made to be associated with three different texts, highlighting a more foundational vulnerability. 

The plausible root cause of such adversarial examples and also semantically different images with identical embeddings is that transformers do not require the inputs to be aligned to have similar embeddings. By adding alignment-sensitive components to the embedding could mitigate the problem, which is being investigated further. 

Given that the models are susceptible to such adversarial attacks, a logical question is if there is an effective method to mitigate the attacks. One potential way to do so is to train a model further to reduce the vulnerabilities. For deep neural networks, robust adversarial training has been used with success~\citep{bai2021recent}. It is unclear how much an adversarially trained model will affect the algorithm's ability to match images with text, and this is currently being investigated.

The results shown in this paper seem not to be consistent with the impressive results demonstrated by such models. Note that almost all existing results are measured on benchmark datasets. Due to the high dimensionality of the embedding space and the input space, even the largest dataset will cover the spaces very sparsely. We believe that systematic evaluations such as ours are necessary if one likes to evaluate models to be able to predict their behaviors in the entire space rather than on samples. 

\section{Conclusion}
In this paper, using a gradient descent-based procedure, we have revealed a new vulnerability in multimodal models, where semantically unrelated inputs can have similar representations, and, at the same time, semantically identical images can have very different representations. Therefore, aligning different inputs to shared embedding space in a semantically meaningful way may not be viable. As multiple models are being developed, one must consider the vulnerabilities in multimodal models for secure applications. As the proposed technique can associate any image with any chosen text, one must understand the implications of this inherent vulnerability.

\bibliography{reference}

\clearpage

\appendix

\section{Appendix}
\label{sec:appendix}

\subsection{Vision Transformers}

Very recently, several multi-modal models have been introduced. By using a shared embedding space among different modalities, such joint models have shown to have advantages. Vision transformers have been successful in various vision tasks due to their ability to treat an image as a sequence of patches and utilize self-attention mechanisms.

\begin{figure}[H]
  \centering
    \includegraphics[width=0.90\columnwidth]{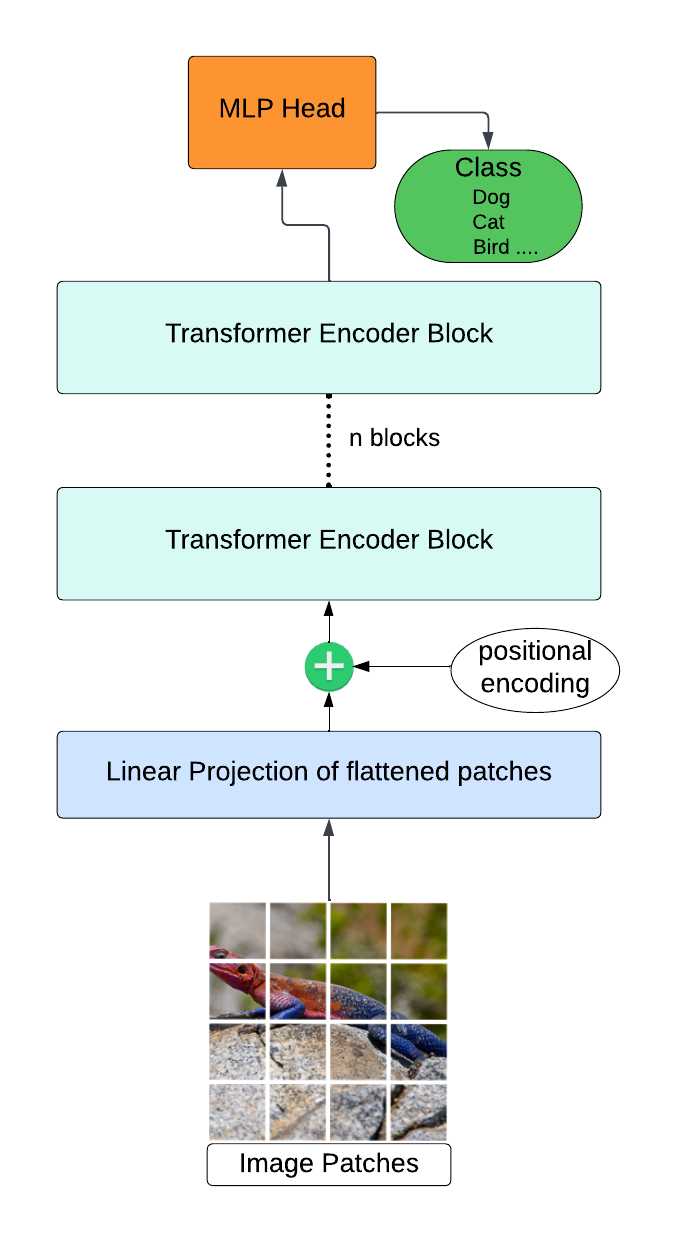}
  \caption{Vision Transformer (ViT) architecture \citep{dosovitskiy2021image}}
  \label{fig:vt_arch}
\end{figure}

A collection of transformer blocks make up the Vision Transformer Architecture. Each transformer block comprises two sub-layers: a multi-headed self-attention layer and a feed-forward layer. The self-attention layer computes attention weights for each pixel in the image based on its relationship with all other pixels, while the feed-forward layer applies a non-linear transformation to the self-attention layer's output. The patch embedding layer separates the image into fixed-size patches before mapping each patch to a high-dimensional vector representation. These patch embeddings are then supplied into the transformer blocks to be processed further \citep{dosovitskiy2021image}.

\subsection{Additional Results}
Here we provide more details and additional information about the results we have included in the main text.

\begin{figure}[ht]
    \centering
\begin{tabular}{>{\centering\arraybackslash}m{.08\textwidth}m{.35in}>{\centering\arraybackslash}m{.09\textwidth}m{.05in}>{\centering\arraybackslash}m{.1\textwidth}}
    \centering\arraybackslash
    \includegraphics[width=.10\textwidth]{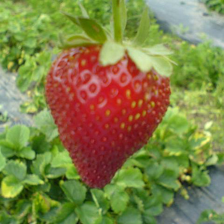} &%
    \centering\arraybackslash%
$\ +\ .04\ \times$ &%
    \includegraphics[width=.10\textwidth]{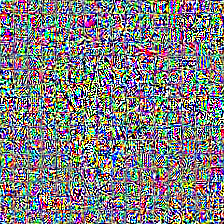} &%
    $=$ & %
    \includegraphics[width=.10\textwidth]{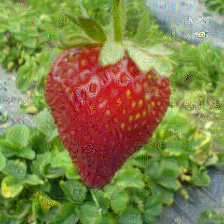} \\

    \centering\arraybackslash
    \includegraphics[width=.10\textwidth]{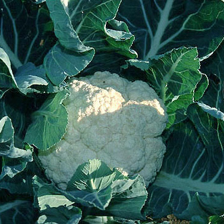} &%
    \centering\arraybackslash%
$\ +\ .04\ \times$ &%
    \includegraphics[width=.10\textwidth]{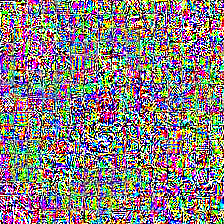} &%
    $=$ & %
    \includegraphics[width=.10\textwidth]{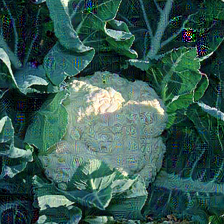} \\
       
\end{tabular}
    \caption{
  Pixel differences between the original and corresponding embedding-aligned images in Fig. \ref{fig:overall} (b) and (f); they are multiplied by 25 for visualization. 
    }
\label{fig:diffnoise}
\end{figure}

\begin{table}
  \centering
    \small
  \begin{tabular}{llll}
    \toprule
    Image & Text & Mean PSNR & Mean SSIM\\
    \midrule
    ImageNet & 1,2,3-tokens & 43 dB & 0.980  \\
    ImageNet & Jigsaw toxic & 47 dB & 0.985 \\
    MS-COCO & 1,2,3-tokens & 46 dB & 0.986 \\
    MS-COCO & Jigsaw toxic & 45 dB & 0.982 \\
    
    \bottomrule
  \end{tabular}
  \label{tab:psnr_ssim}
  \caption{The average PSNR value and SSIM index between the original and embedding-aligned images to all the text embeddings in each of the datasets; the average is computed based on 800 examples for each dataset and the images and texts are strictly randomly chosen from the datasets with no postselection.
  }
\end{table}

\textbf{Image Quality Evaluation. } 
Peak Signal-to-Noise Ratio (PSNR) and Structural Similarity Index (SSIM) are commonly used metrics to quantify the differences between the original and modified images~\cite{alain2010psnr, morales2019dehaze}. PSNR effectively measures the detailed quality of an image, whereas SSIM provides an intuitive assessment of its structural integrity. We present the average PSNR and SSIM values between the original and manipulated (i.e., embedding-aligned) images across all the datasets under consideration in Table 2. These metrics indicate that the image quality does not significantly degrade with minimal distortion. Due to resource and time constraints, we restricted the results to 800 examples for Table 2. We followed the approach by Szegedy et al.~\cite{szegedy2014intriguing}, where they used a smaller set (64 images) from ImageNet when calculating the average distortion of adversarial examples.

\textbf{More Results. }In the main paper, the results are mostly generated using the ImageNet and 1,2,3-tokens toxic dataset. To showcase the versatility of our framework across different vision and text datasets, the subsequent figures also present results obtained from additional datasets such as MS-COCO and Jigsaw toxic. 

\begin{figure*}[ht]
    \centering
      \includegraphics[width=0.98\textwidth]{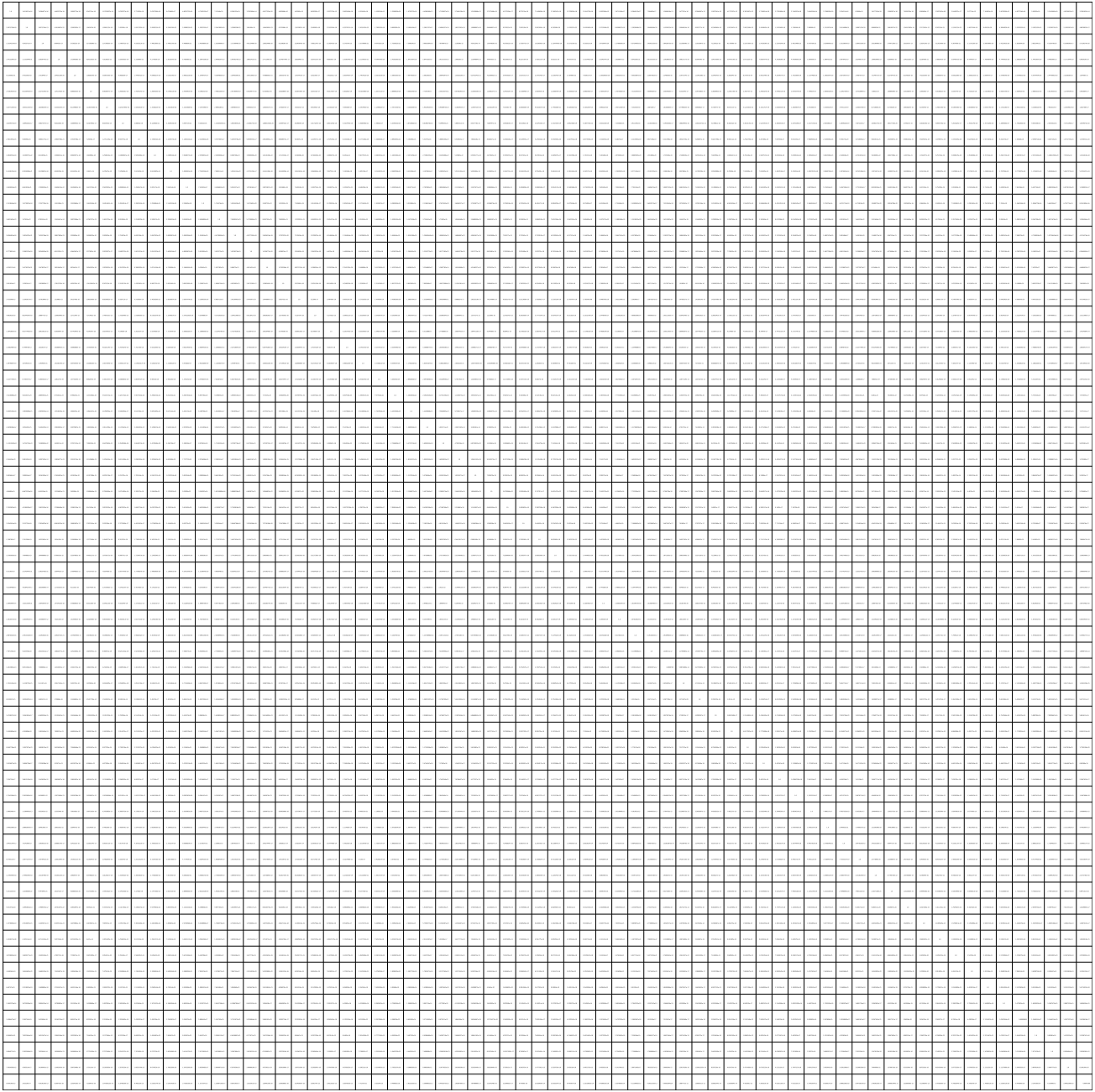}\label{fig:more_ex1}
    \caption{(For optimal viewing in PDF, zoom in) The original unclipped outputs from ImageBind model when an ImageNet example (e.g., strawberry) aligns with all 68 comments from the 1-token toxic dataset, achieving a 100\% match success rate. Please zoom in to see the classification probabilities more clearly.}
 \label{fig:unclipped_68}
\end{figure*}

Figure \ref{fig:unclipped_68} shows the original outputs from the joint
vision × text ImageBind model when an ImageNet example matches with all the
68 comments of the 1-token toxic dataset, therefore getting 100\% match success rate. It is clear that values are either very close to 1 or very close to 0, demonstrating that the classification results are stable.

\begin{figure*}[ht]
    \centering
      \includegraphics[width=0.950\textwidth]{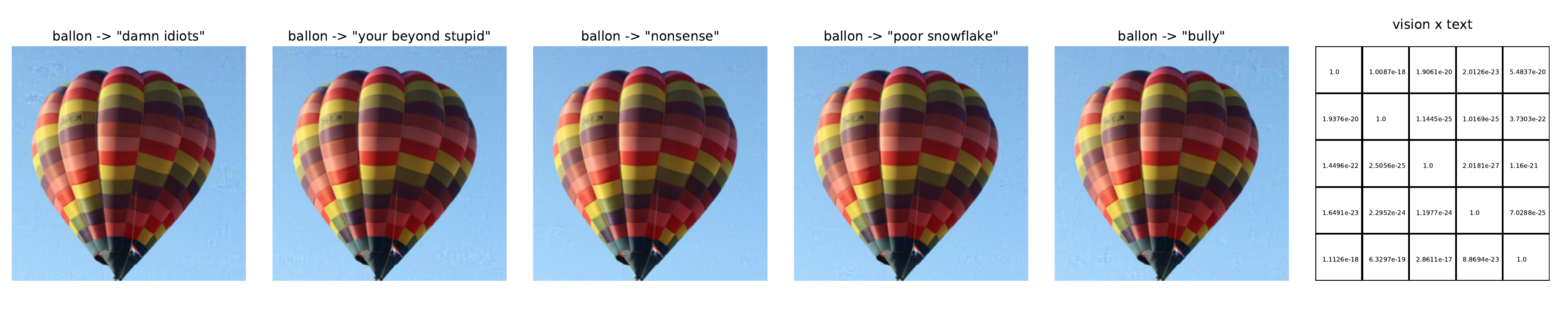}\label{fig:more4}
    
      \includegraphics[width=0.950\textwidth]{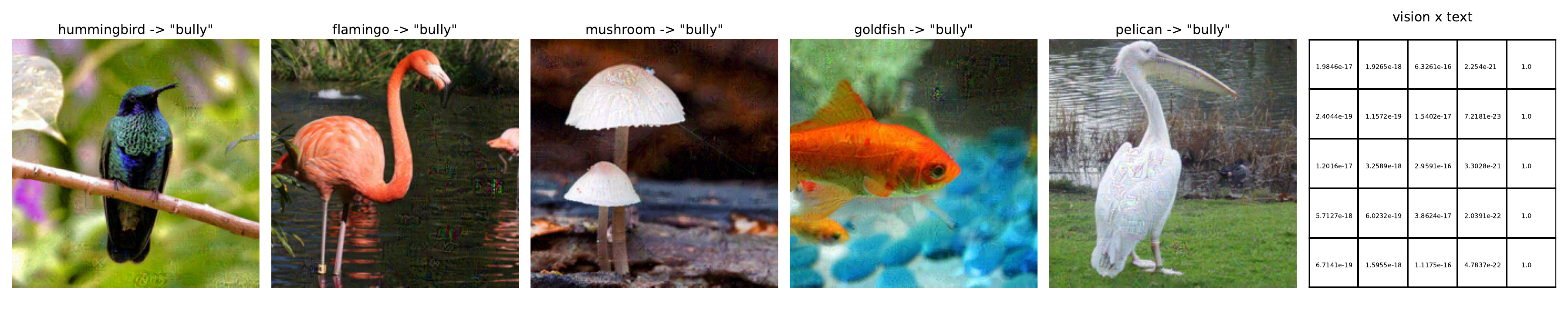}\label{fig:more5}

    \caption{Additional examples involving ImageNet and 1,2,3-tokens dataset. (top) Visually indistinguishable images have very different representations via embedding alignment with the corresponding texts and therefore very different classification outcomes. (bottom) Visually very different images have very similar embeddings, aligned and classified to a particular text. The examples are strictly randomly chosen. There is no postselection involved.}
 \label{fig:more_examples_3}
\end{figure*}

\begin{figure*}[ht]
    \centering
      \includegraphics[width=0.950\textwidth]{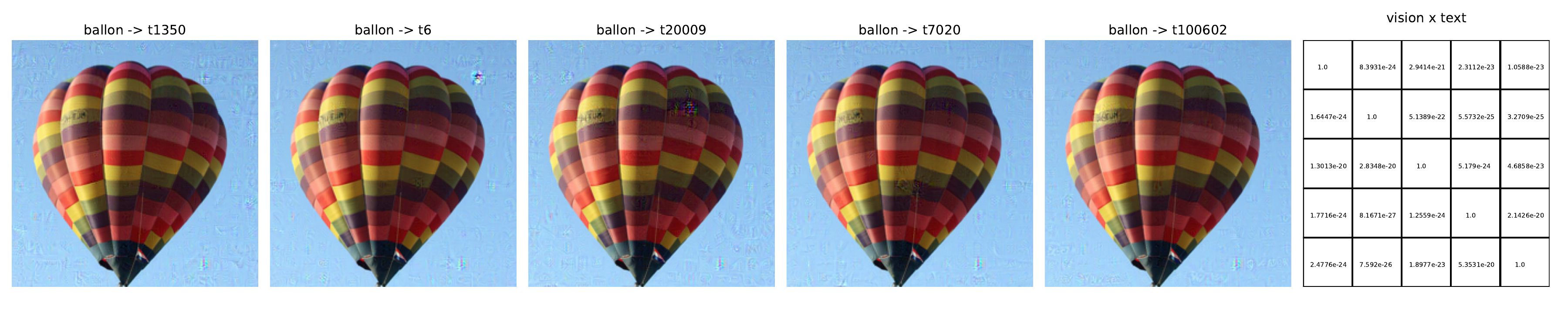}\label{fig:more6}
    
      \includegraphics[width=0.950\textwidth]{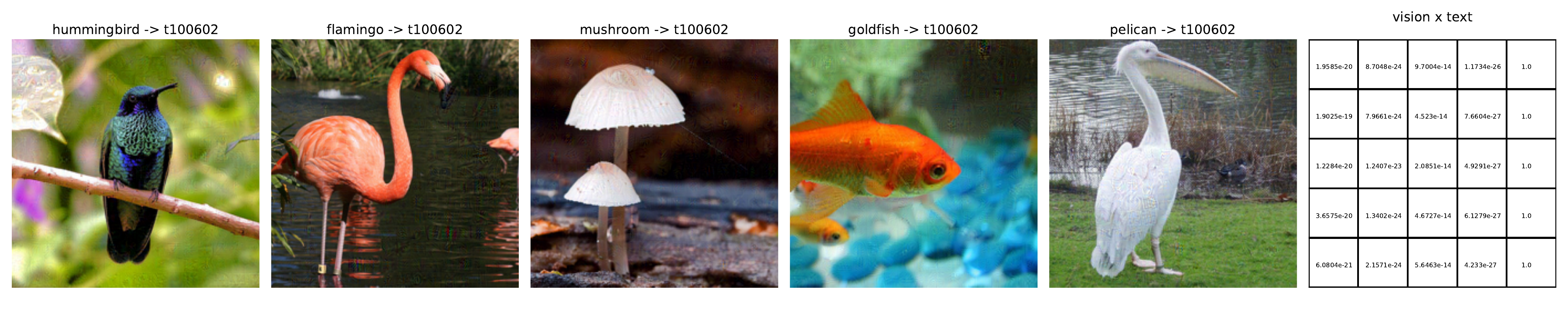}\label{fig:more7}

    \caption{More examples involving ImageNet and Jigsaw toxic dataset. (top) Visually indistinguishable images have very different representations via embedding alignment with the corresponding texts and therefore very different classification outcomes. (bottom) Visually very different images have very similar embeddings, aligned and classified to a particular text.}
 \label{fig:more_examples_4}
\end{figure*}

\begin{figure*}[ht]
    \centering
      \includegraphics[width=0.950\textwidth]{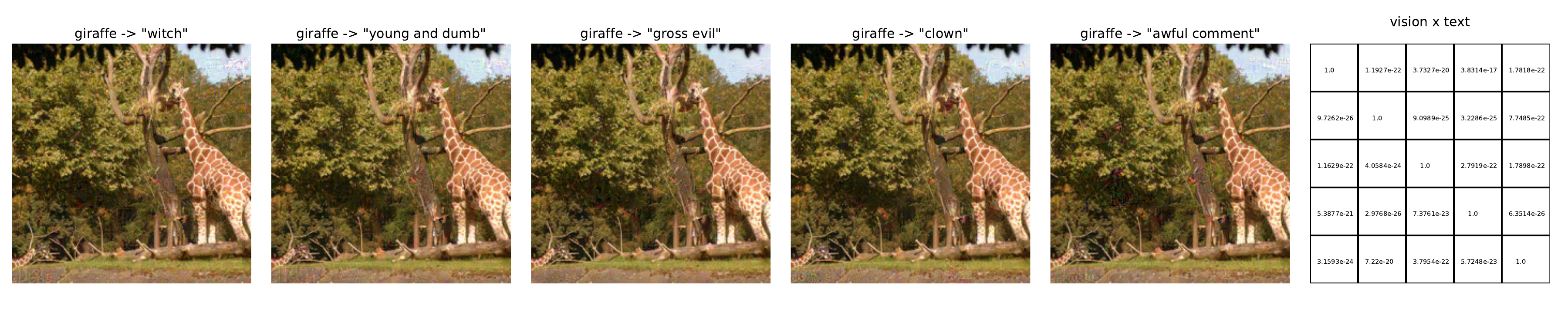}\label{fig:more8}
    
      \includegraphics[width=0.950\textwidth]{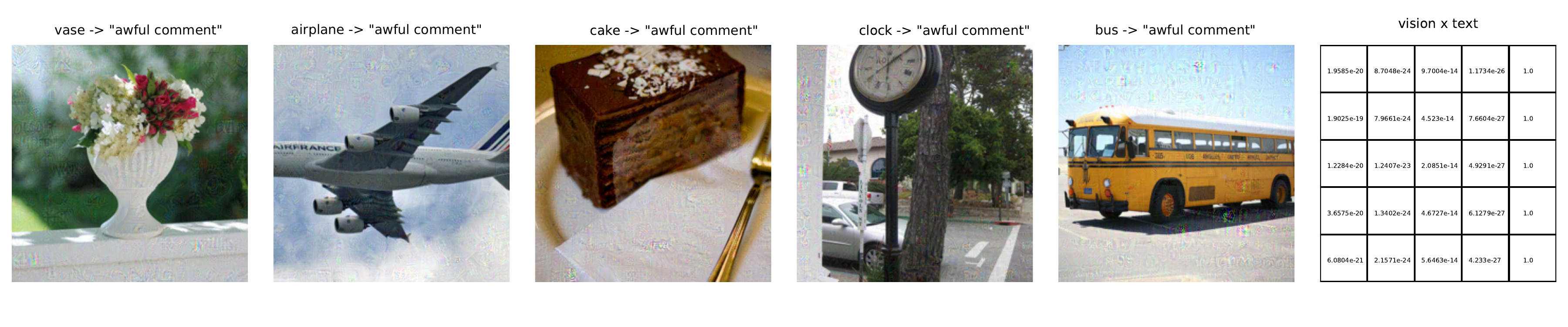}\label{fig:more9}

    \caption{Additional examples involving MS-COCO and 1,2,3-tokens toxic dataset. (top) Visually indistinguishable images have very different representations via embedding alignment with the corresponding texts and therefore very different classification outcomes. (bottom) Visually very different images have very similar embeddings, aligned and classified to a particular text. Again the samples are randomly chosen.}
 \label{fig:more_examples_5}
\end{figure*}

\begin{figure*}[ht]
  \centering
  \includegraphics[width=0.95\textwidth]{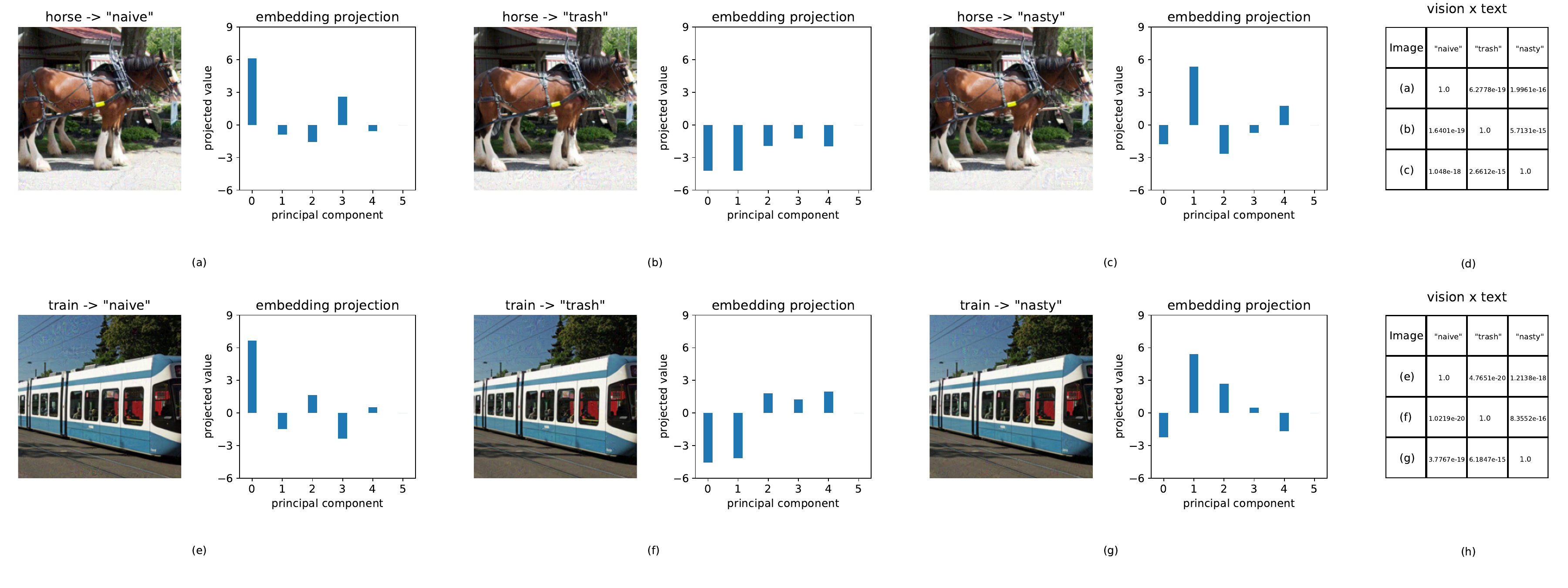}\label{fig:projection_embeddings_ht}
  \caption{Same as Fig. \ref{fig:overall}, but shown while the proposed framework is applied on MS-COCO data examples and 1,2,3-tokens toxic comments. Again the samples are randomly chosen.
}
  \label{fig:overall_5}
\end{figure*}

\end{document}